\documentclass[11pt,a4paper]{article}
\usepackage[hyperref]{emnlp2020}
\usepackage{times}
\usepackage{latexsym}

\usepackage{microtype}

\aclfinalcopy 

\usepackage{booktabs}
\usepackage{multirow}
\usepackage[utf8]{inputenc}
\usepackage[english,russian,greek,arabic]{babel}
\usepackage[export]{adjustbox}

\newcommand{\good}[1]{\underline{\textcolor{cyan}{#1}}}
\newcommand{\bad}[1]{\textit{\textcolor{red}{#1}}}

\usepackage{subfigure}
\usepackage{graphicx}
\usepackage[normalem]{ulem}
\title{It's not Greek to mBERT:\\ Inducing Word-Level Translations from Multilingual BERT}

\author{Hila Gonen\textsuperscript{1} \, Shauli Ravfogel\textsuperscript{1,2} \, Yanai Elazar\textsuperscript{1,2} \, Yoav Goldberg\textsuperscript{1,2}\\
\textsuperscript{1}Computer Science Department, Bar Ilan University \\
\textsuperscript{2}Allen Institute for Artificial Intelligence \\
  {\tt  \{hilagnn, shauli.ravfogel, yanaiela, yoav.goldberg\}@gmail.com}
  }

\date{}

\begin{document}

\selectlanguage{english}

\maketitle

\begin{abstract}

Recent works have demonstrated that multilingual BERT (mBERT) learns rich cross-lingual representations, that allow for transfer across languages. We study the word-level translation information embedded in mBERT and present two simple methods that expose remarkable translation capabilities with no fine-tuning. The results suggest that most of this information is encoded in a non-linear way, while some of it can also be recovered with purely linear tools. 
As part of our analysis, we test the hypothesis that mBERT learns representations which contain both a language-encoding component and an abstract, cross-lingual component, and explicitly identify an empirical language-identity subspace within mBERT representations.

\end{abstract}

\section{Introduction}

Multilingual-BERT (mBERT) is a version of BERT \cite{DCL19}, trained on the concatenation of Wikipedia in 104 different languages. Recent works show that it excels in zero-shot transfer between languages, for a variety of tasks \cite{PSG19,MSS20}, despite being trained with no parallel supervision.

Previous work has mainly focused on what is needed for zero-shot transfer to work well \cite{MSS20,KWM20,WD19}, and on characterizing the representations of mBERT \cite{SMS19}. However, we still lack a proper understanding of this model.

In this work we study (1) how much word-level translation information is recoverable by mBERT; and (2) how this information is stored. 
We focus on the representations of the last layer, and on the embedding matrix that is shared between the input and output layers -- which are together responsible for token prediction.

For our first goal, we start by presenting a simple and strong method to extract word-level translation information. Our method is based on explicit querying of mBERT: given a source word and a target language, we feed mBERT with a template such as \textit{``The word `\textsc{source}' in \textsc{language} is: [MASK]."} where \textsc{language} is the target language, and \textsc{source} is an English word to translate. 
Getting the correct translation as the prediction of the masked token exposes mBERT's ability to provide word-level translation. This template-based method is surprisingly successful, especially considering the fact that no parallel supervision was provided to the model while training, and that word translation is not part of the training objective.

This raises the possibility of easy disentanglement between language identity and lexical semantics in mBERT representations. 
We test this hypothesis by trying to explicitly disentangle language-identity from lexical semantics under linearity assumptions. We propose a method for disentangling a language-encoding component and a language-neutral component from both the embedding representations and word-in-context representations. Furthermore, we learn the emperical ``langauge subspace" in mBERT, which is a linear subspace that is spanned by all directions that are linearly correlative to the language identity. We demonstrate that the representations are well-separated by language on that subspace.

We leverage these insights and empirical results to show that it is possible to perform analogies-based translation by taking advantage of this disentanglement: we can alter the language-encoding component, while keeping the lexical component intact. We compare between the template-based method and the analogies-based method and discuss their similarities and differences, as well as their limitations.

The two methods together show that mBERT acquired, to a large degree, the ability to perform word-level translation, despite the fact that it is not trained on any parallel data explicitly. The results suggest that most of the information is stored in a non-linear way, but with some linearly-recoverable components. 

Our contribution in this work is two-fold: (a) we present two simple methods for word-level translation using mBERT, that require no training or finetuning of the model, which demonstrate that mBERT stores parallel information in different languages; (b) we show that mBERT representations are composed of language-encoding and language-neutral components and present a method for extracting those components. Our code is available at \url{https://github.com/gonenhila/mbert}.

\section{Previous Work}

\newcite{PSG19} begin a line of work that studies mBERT representations and capabilities. In their work, they inspect the model's zero-shot transfer abilities using different probing experiments, and propose a way to map sentence representations in different languages, with some success. \newcite{KWM20} further analyze the properties that affect zero shot transfer by experimenting with bilingual BERTs on RTE (recognizing textual entailment) and NER. They analyse performance with respect to linguistic properties and similarities of the source and the target languages, and some parameters of the model itself (e.g. network architecture and learning objective). In a closely related work, \newcite{WD19} perform transfer learning from English to 38 languages, on 5 tasks (POS, parsing, NLI, NER, Document classification), and report good results. Additionally, they show that language-specific information is preserved in all layers. \newcite{WCG19} learn alignment between contextualized representations, and use it for zero shot transfer.

Beyond focusing on zero-shot transfer abilities, an additional line of work studies the representations of mBERT and the information it stores. Using hierarchical clustering based on the CCA similarity scores between languages, \newcite{SMS19} are able to construct a tree structure that faithfully describes relations between languages. \citet{mbert-syntax} learn a linear syntax-subspace in mBERT, and point out to syntactic regulartieis in the representations that transfer across languages.
In the recent work of \newcite{CKK20}, the authors define the notion of \textit{contextual} word alignment. They design a fine-tuning loss for improving alignments and show that they are able to improve zero-shot transfer after this alignment-based fine-tuning. One main difference from our work is that they fine-tune the model according to their new definition of contextual alignment, while we analyze and use the information already stored in the model.
One of the closest works to ours is that of \newcite{LRF19}, where they assume that mBERT's representations have a language-neutral component, and a language-specific component. They remove the language specific component by subtracting the centroid of the language from the representations, and make an attempt to prove the assumption by using probing tasks on the original vs. new representations. They show that the new representations are more language-neutral to some extent, but lack experiments that show a complementary component. 
While those works demonstrate that mBERT representations in different languages can be aligned successfully with appropriate supervision, we propose an explicit \emph{decomposition} of the representations to language-encoding and language-neutral components, and also demonstrate that implicit word-level translations can be easily distilled from the model when exposed to the proper stimuli.

\section{Word-level Translation using Pre-defined Templates}

We study the extent to which it is possible to extract word-level translation directly from mBERT.

\subsection{Word-level Translation: You Just Have to Ask}

We present a simple and overwhelmingly successful method for word-level translation with mBERT. This method is based on the idea of explicitly querying mBERT for a translation, similar to what has been done with LMs for other tasks \cite{petroni2019language,jiang2019can,talmor2019olmpics}. 
We experimented with seven different templates and found the following to work best:
\textit{``The word `\textsc{source}' in \textsc{language} is: [MASK]."}\footnote{where \textsc{source} is the word we wish to translate and [MASK] is the special token that BERT uses as an indication for word prediction, see Section~\ref{templates_app} in the Appendix for the other templates.}
The predictions from the [MASK] token induce a distribution over the vocabulary, and we take the most probable word as the translation.

\begin{table}[h!]
    \centering
    \resizebox{\columnwidth}{!}{
    \begin{tabular}{l|lll|ll|l}
    & @1 & @10 & @100 & rank & log & win \\ \hline \hline
    
    Baseline & 0.036 &  0.244 & 0.575 & 4303.4 & 4.58 & -- \\
    \hline \hline
    Analogies & 0.105 & 0.463 & 0.737 & 2458.1 & 3.19 & 89.8\% \\ 
     \hline 
    Template & \textbf{0.449} & \textbf{0.703} & \textbf{0.845}  & \textbf{243.4} & \textbf{1.68} & \textbf{91.6}\% \\ \hline 
    
    \end{tabular}}
    \caption{Word-level translation results with the template-based method and the analogies-based method (introduced in Section~\ref{analogies}). @1-100 stand for accuracy@$k$ (higher is better), ``rank" stands for the average rank of the correct translation, ``log" stands for the log of the average rank, and ``win" stands for the percentage of cases in which the tested method is strictly better than the baseline.}
    \label{tab:acc}
\end{table}

\subsection{Evaluation}

To evaluate lexical translation quality, we use NorthEuraLex\footnote{\url{http://northeuralex.org/}} \citep{DDM19}, a lexical database providing translations of 1016 words into 107 languages. We use these parallel translations to evaluate our translation method when translating from English to other target languages.\footnote{We use this data also for experimenting with source languages other than English, in Section~\ref{analogies}.} We restrict our evaluation to a set of common languages from diverse language families: Russian, French, Italian, Dutch, Spanish, Hebrew, Turkish, Romanian, Korean, Arabic and Japanese. We omit cases in which the source word or the target word are tokenized using mBERT into more than a single token.\footnote{Number of translated pairs we are left with in each language: Russian: 224, French: 429, Italian: 352, Dutch: 347, Spanish: 452, Hebrew: 158, Turkish: 199, Romanian: 243, Korean: 42, Arabic: 191, Japanese: 214.} The words in the dataset are from different POS, with Nouns, Adjectives and Verbs being the most common ones.\footnote{Number of words from each POS: `N': 480, `V': 340, `A': 102, `ADV': 47, `NUM': 22, `PRN': 9, `PRP': 7, `FADV': 4, `FPRN': 2, `CNJ': 2, `FNUM': 1.}
For all our experiments with mBERT, we use the transformer library of HuggingFace \cite{trans_hf}.

\paragraph{Results}

We report accuracy@$k$ in translating the English source word into different languages (for $k \in \{1,10,100\}$): for each word pair, we check whether the target word was included in the first $k$ retrieved words. Note that we remove the source word itself from the ranking.\footnote{This removal mainly affects acc@1, since in many cases, the first retrieved word is the source word. This is a common practice, especially for the analogies-based method~\cite{mikolov2013}.} We report three additional metrics: (a) \textbf{avg-rank}: the average rank of the target word (its position in the ranking of predictions); (b) \textbf{avg-log-rank}: the average of the log of the rank, to limit the effect of cases in which the rank is extremely low and skews the average; (c) \textbf{hard-win}: percent of cases in which the method results in a strictly better rank for the translated word compared to the baseline.
We take the predictions we get for the masked token as the method's candidates for translation. As a baseline, we take the embedding representation of the source word and look for the closest words to it. 
Table~\ref{tab:acc} shows the results of the template-based method and the baseline. This method significantly improves over the baseline in all metrics and achieves impressive accuracy results: acc@1 of 0.449 and acc@10 of 0.703, beating the baseline in 91.6\% of the cases.

\paragraph{Accuracy per POS}

To get a finer analysis of this method, we also evaluate the translations per POS. We report results on the 3 most common POS: nouns, adjectives and verbs.\footnote{We have 1765, 363, 323 instances, respectively. Other POS have less than 200 instances, and are thus omitted from the analysis.} As one might expect, nouns are the easiest to translate (both for the baseline and for our method), followed by adjectives, then verbs. See Table~\ref{tab:acc-pos} for full results.

Note that the results for these common POS tags are lower than the average over the full dataset. We hypothesize that words belonging to closed-class POS tags, such as pronouns, are easier to translate.

\begin{table}
    \centering
    \scalebox{0.8}{
    \begin{tabular}{l|l|l|l|l|l}
    & @1 & @5 & @10 & @50 & @100 \\ \hline \hline
    \multirow{2}{*}{Noun} & \small{(0.044)} & \small{(0.189)} & \small{(0.279)} & \small{(0.520)} & \small{(0.593)} \\
    & 0.113 & 0.379 & 0.494 & 0.696 & 0.741    \\ \hline 
    \multirow{2}{*}{Adjective} & \small{(0.037)} & \small{(0.102)} & \small{(0.183)} & \small{(0.474)} & \small{(0.567)} \\
    & 0.105 & 0.310 & 0.418 & 0.666 & 0.706     \\ \hline
    \multirow{2}{*}{Verb} & \small{(0.006)} & \small{(0.105)} & \small{(0.168)} & \small{(0.355)} & \small{(0.427)} \\
    & 0.039 & 0.201 & 0.317 & 0.576 & 0.645     \\ \hline
    \end{tabular}}
    \caption{Word-level translation results per POS with the template-based method. The numbers in parathesis relate to the baseline.}
    \label{tab:acc-pos}
\end{table}

\subsection{Visualization of the Representation Space}

\begin{figure}
	\centering
	\scalebox{0.4}{
	\includegraphics{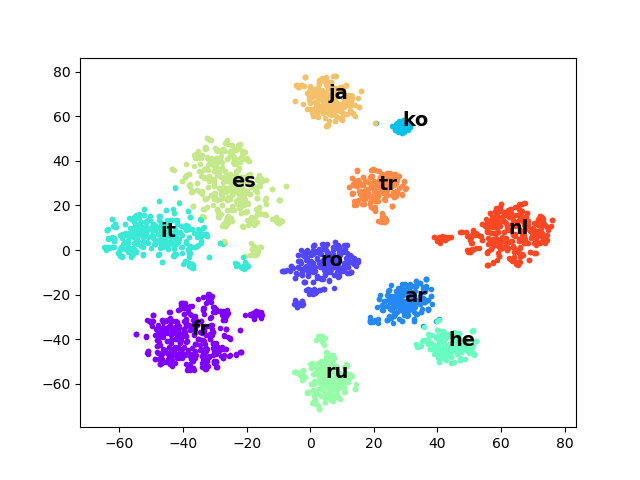}}
	\caption{t-SNE projections of the representations of the template-based method.}
	\label{fig:tsne}
\end{figure}

To further understand the mechanism of the method, we turn to inspect the resulting representations. For each word pair, we feed mBERT with the full template and extract the last-layer representation of the masked token, right before the multiplication with the output embeddings. In Figure~\ref{fig:tsne} we plot the t-SNE projection \citep{tsne} of those representations, colored by language. The representations clearly cluster according to the target language. The ability of these representations to encode the target language may explain how this method successfully produces the translation into the correct language.

\subsection{Predicting the Language}
\label{sec:predict_lang}

Due to the representations clustering based on the target language (rather than semantics), we hypothesize that mBERT is also capable of predicting the target language given the source word and its translation.

To verify that, we take the same template as before, this time masking the name of the language instead of the target word.\footnote{We use all languages in NortEuraLex that are a single token according to mBERT tokenization -- there are 47 such languages except for English.} We then compute acc@1,5,10 for all languages and report that for the 20 languages with the most accurate results in Table~\ref{tab:lang_guess} (the full results can be found in Table~\ref{tab:lang_guess_app} in the Appendix). The results are impressive, suggesting that mBERT indeed encodes the target language identity in this setting. The languages on which mBERT is most accurate are either widely-spoken languages (e.g. German, French), or languages with a unique script (e.g. Greek, Russian, Arabic). Indeed, we get a Spearman correlation of 0.53 between acc@1 and the amount of training data in each language.\footnote{We considered the number of articles per language from Wikipedia: \url{https://en.wikipedia.org/wiki/Wikipedia:Multilingual_statistics}, recorded in May 2020.}

\begin{figure}
	\centering
	
	\scalebox{0.2}{
	\includegraphics{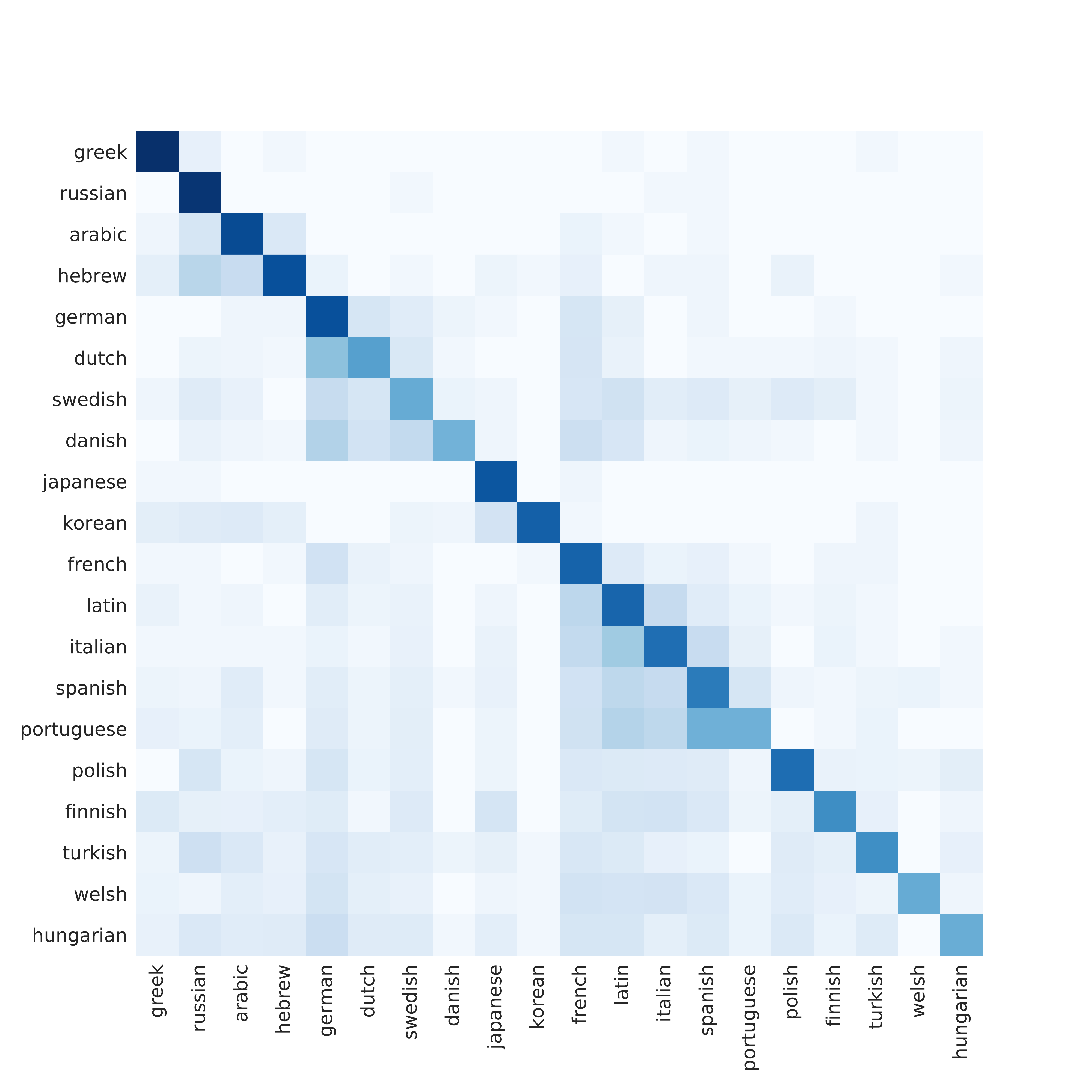}}
	\caption{Confusion matrix of language prediction when the language is masked in the template. 20 most accurate languages are included, English is omitted. }
	\label{fig:confusion}
\end{figure}

We also compute a confusion matrix for the 20 most accurate languages, shown in Figure~\ref{fig:confusion}. In order to better identify the nuances, we use the square-root of the values, instead of the values themselves, and remove English (which is frequently predicted as the target language, probably since the template is in English). The confusion matrix reveals the expected behavior -- mBERT confuses mainly between typologically related languages, specifically those of the same language family: Germanic languages (German, Dutch, Swedish, Danish), Romance languages (French, Latin, Italian, Spanish, Portuguese), and Semitic languages (Arabic, Hebrew). In addition, we can also identify some confusion between Germanic and Romance languages (which share much of the alphabet), as well as over-prediction of languages with a lot of training data (e.g. German, French).

\begin{table}[t!]
    \centering
    \scalebox{0.8}{
    \begin{tabular}{l|l|l|l}
    Language & acc@1 & acc@5 & acc@10 \\ \hline \hline
    greek & 0.986 & 0.998 & 1.000 \\ \hline
    russian & 0.943 & 0.994 & 0.998 \\ \hline
    arabic & 0.794 & 0.963 & 0.984 \\ \hline
    hebrew & 0.761 & 0.968 & 0.990 \\ \hline
    german & 0.758 & 0.951 & 0.991 \\ \hline
    japanese & 0.716 & 0.939 & 0.966 \\ \hline
    korean & 0.664 & 0.905 & 0.949 \\ \hline
    french & 0.637 & 0.976 & 0.992 \\ \hline
    latin & 0.626 & 0.900 & 0.959 \\ \hline
    polish & 0.576 & 0.728 & 0.803 \\ \hline
    italian & 0.572 & 0.873 & 0.947 \\ \hline
    spanish & 0.503 & 0.757 & 0.878 \\ \hline
    finnish & 0.404 & 0.622 & 0.748 \\ \hline
    turkish & 0.399 & 0.589 & 0.709 \\ \hline
    dutch & 0.315 & 0.846 & 0.965 \\ \hline
    welsh & 0.262 & 0.548 & 0.680 \\ \hline
    swedish & 0.262 & 0.492 & 0.692 \\ \hline
    hungarian & 0.254 & 0.395 & 0.493 \\ \hline
    portuguese & 0.236 & 0.587 & 0.808 \\ \hline
    danish & 0.231 & 0.388 & 0.567 \\ \hline

    \end{tabular}}
    \caption{Prediction accuracy of the language, when the language is masked in the template (20 most accurate languages).}
    \label{tab:lang_guess}
\end{table}

\section{Dissecting mBERT Representations}
\label{sec:components}

In the previous section, we saw that mBERT contains abundant word-level translation knowledge. How is this knowledge represented? We turn to analyze both the representations of words in context and those of the output embeddings.

It has been assumed in previous work that the representations are composed of a language-encoding component and a language-neutral component \cite{LRF19}. In what follows, we explicitly try to find such a decomposition: we decompose $v = v_{lang} + v_{lex}$, where $v_{lang}$ and $v_{lex}$ are orthogonal vectors, $v_{lang}$ is the part in the representation that is indicative of language identity, and $v_{lex}$ maintains lexical information, but is invariant to language identity. Specifically, we test the hypothesis using the following interventions: 

\begin{itemize}
    \item Measuring the degree to which removing $v_{lang}$ results in language-neutral word representations.
    \item Measuring the degree to which removing $v_{lex}$ results in word representations which are clustered by language identity (regardless of lexical semantics).
    \item Removing the $v_{lang}$ component from word-in-context representations and from the output embeddings, to induce MLM prediction in other languages.
\end{itemize}

Splitting the representations into components is done using INLP \cite{inlp}, an algorithm for removing information from vector representations.

\subsection{mBERT Decomposition by Nullspace Projections}
\label{inlp}

We formalize the decomposition objective defined earlier as finding two linear \emph{subspaces} within the representation space, which contain language-independent and language-identity features. The recently proposed Iterative Null-space Projection (INLP) method \citep{inlp} allows to remove linearly-decodable information from vector representations. Given a dataset of representations $X$  (in our case, mBERT word-in-context representations and output embeddings) and annotations $Z$ for the information to be removed (language identity) the method renders $Z$ linearly unpredictable from $X$. It does so by iteratively training linear predictors $w_1, \dots, w_n$ of $Z$, calculating the projection matrix onto their nullspace $P_{N} := P_N(w_1), \dots,  P_N(w_n)$, and transforming $X \gets P_{N}X$. Recall that by the nullsapce definition this guarantees $w_iP_{N}X=0, \forall w_i$, i.e., the features $w_i$ uses for language prediction are neutralized.

While the nullsapce $N(w_1, \dots, w_n)$ is a subspace in which $Z$ is not linearly predictable, the complement rowspace $R(w_1, \dots, w_n)$ is a subspace of the representation space $X$ that corresponds to the property $Z$. In our case, this subspace is \emph{mBERT language-identity subspace}. In the following sections we utilize INLP in two complementary ways: (1) we use the null-space projection matrix $P_N$ to zero out the language identity subspace, in order to render the representations \emph{invariant} to language identity\footnote{to the extent that language identity is indeed encoded in a linear subspace, and that INLP finds this subspace.}; and (2) we use the \emph{rowspace} projection matrix $P_R = I - P_N$ to project mBERT representations onto the language-identity subspace, keeping only the parts that are useful for language-identity prediction. We hypothesize that the first operation would render the representations more language-neutral, while the latter would discard the components that are shared across languages.

\paragraph{Setup}

We start by applying INLP on random representations and getting the two mentioned projection matrices: on the nullspace, and on the rowspace. We repeat this process twice: first, for representations in context, and second, for output embeddings. For each of these two cases, we sample random tokens from 5000 sentences\footnote{For the output embeddings, we exclude tokens that start with ``\#\#", for the last layer representations, sampled tokens may include ``CLS" or ``SEP". } in 15 different languages, extract their respective representations (in context or simply output embeddings), and run INLP on those representations with the objective of identifying the language, for 20 iterations. We end up with 4 matrices: projection matrix on the null-space and on the rowspace for representations in context, and the same for output embeddings.

\paragraph{TED corpus}

For the experiments depicted in Sections \ref{sec:components} and \ref{analogies}, we use a dataset of transcripts of TED talks in 60 languages, collected by \citet{ted}\footnote{\url{https://github.com/neulab/word-embeddings-for-nmt}}. For the INLP trainings, we use the 15 most frequent languages in the dataset after basic filtering, 12 of which are also included in NorthEuraLex.

\subsection{Language-Neutral and Language-Encoding Representations}

We aim to use INLP nullspace and rowspace projection matrices as an intervention that is designed to test the hypothesis on the exsitence of two independent subspaces in mBERT. Concretely, we perform two experiments: (a)~a cluster analysis, using t-SNE \citep{tsne} and a cluster-coherence measure, of representations projected on the null-space and the row-space from different languages. We expect to see decreased and increased separation by language identity, respectively; (b)~we perform nullsapce projection intervention on both the last hidden state of mBERT, and on the output embeddings, and proceed to predict a distribution over all tokens. We expect that neutralizing the language-identity information this way will encourage mBERT to perform semantically-adequate word prediction, while decreasing its ability to choose the correct language in the context of the input sentence.

\paragraph{t-SNE and Clustering}

\begin{figure}[t!]
\subfigure{
\includegraphics[width=\columnwidth]{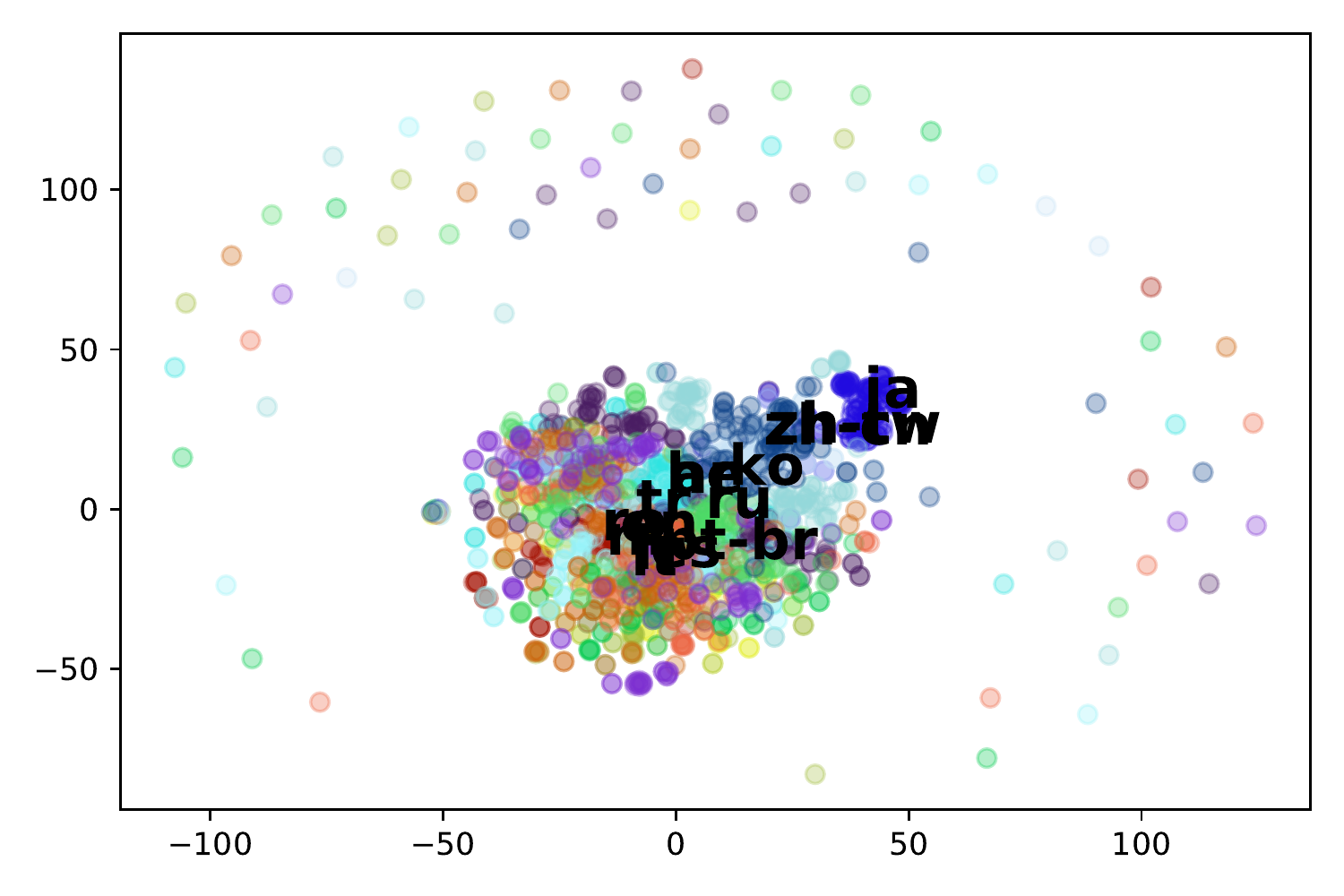}
}
\subfigure{
\includegraphics[width=\columnwidth]{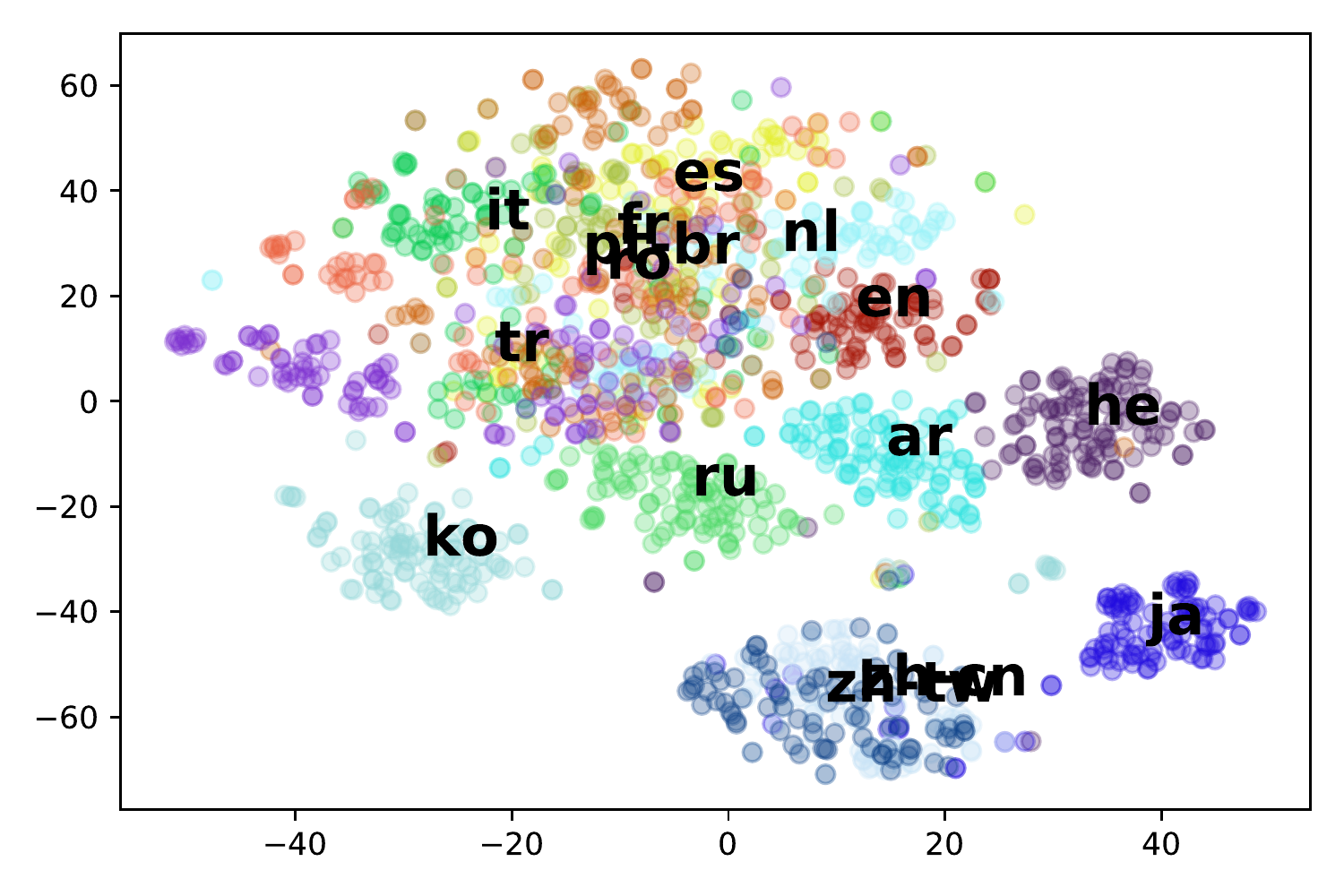}
}
\caption{t-SNE projection of the output embeddings of random words from different languages, originally (top) and after projection onto the language-identity subspace (bottom).
}
\label{fig:tsne-original-vs-rowspace:layer0}
\end{figure}

\begin{figure}[t!]
\subfigure{
\includegraphics[width=\columnwidth]{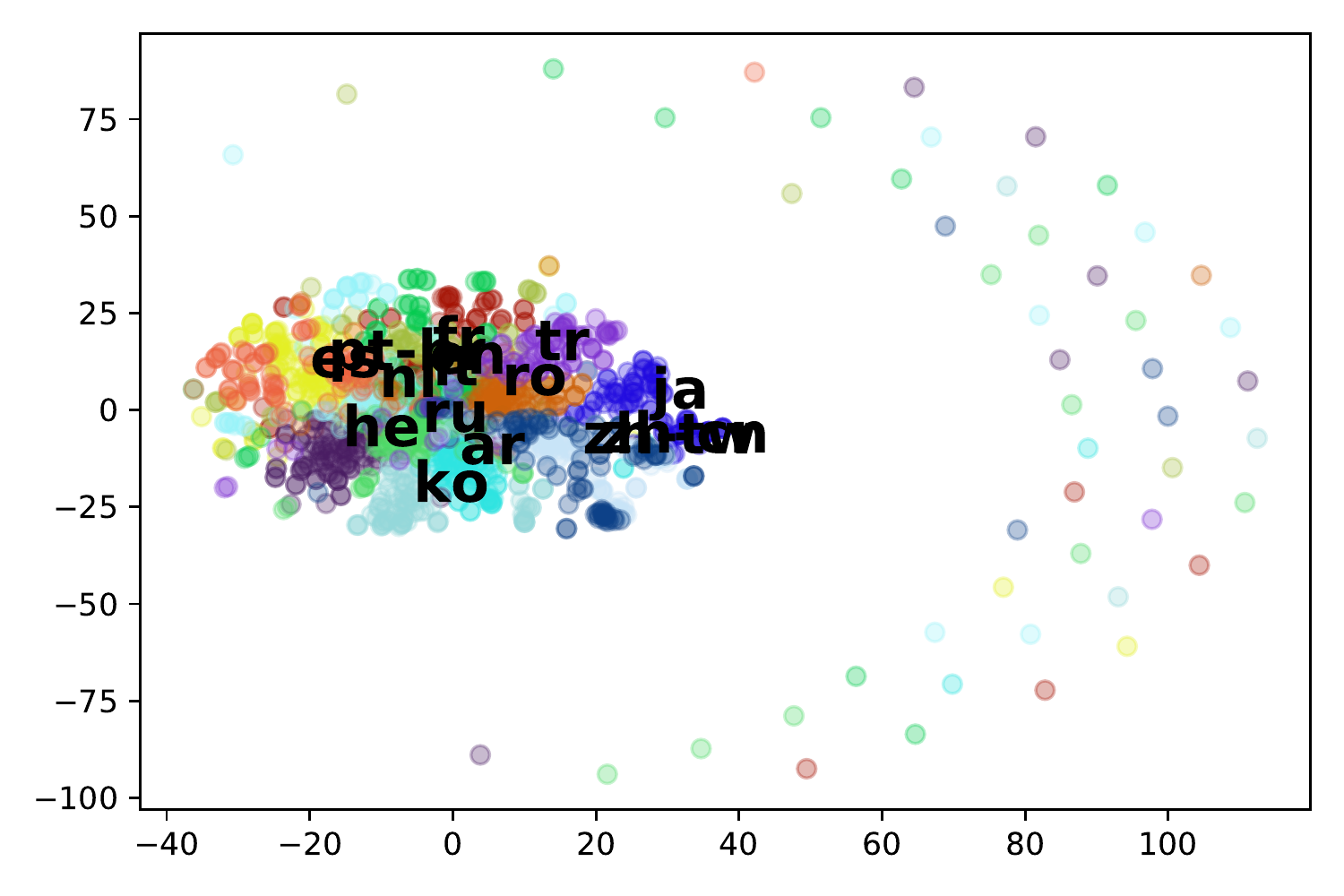}
}
\subfigure{
\includegraphics[width=\columnwidth]{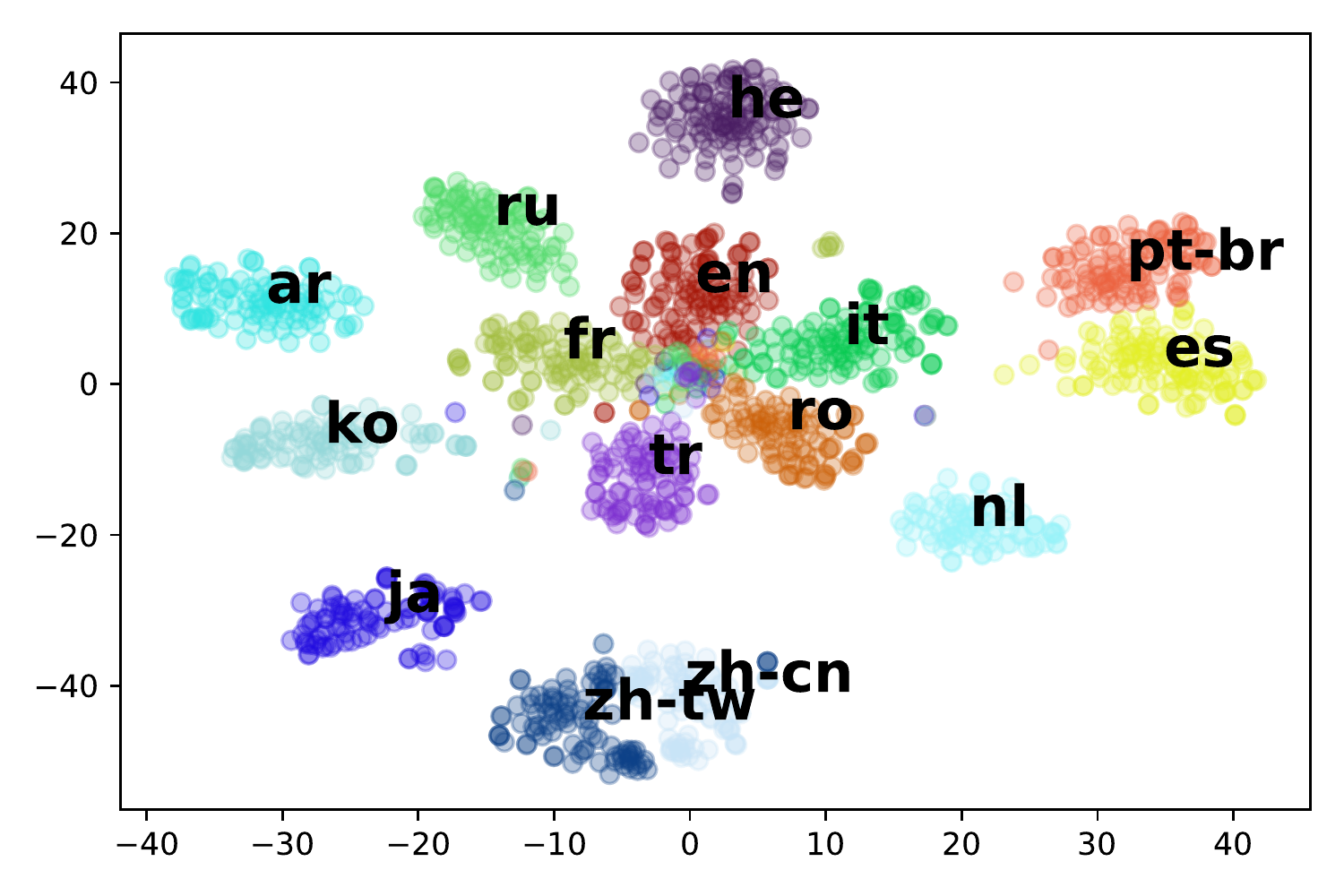}
}
\caption{t-SNE projection of last-hidden-layer representation of random words from different languages, originally (top) and after projection onto the language-identity subspace (bottom).}
\label{fig:tsne-original-vs-rowspace:layer12}
\end{figure}

To test the hypothesis on the existence of a ``language-identity" subspace in mBERT, we project the representations of a random subset of words from TED dataset, from the embedding layer and the last layer, on the subspace that is spanned by all language classifiers, using INLP rowspace-projection matrix. Figures \ref{fig:tsne-original-vs-rowspace:layer0} and \ref{fig:tsne-original-vs-rowspace:layer12} present the results for the embedding layer and the last layer, respectively. In both cases, we witness a significant improvement in clustering according to language identity. At the same time, some different trends are observed across layers: the separability is better in the last layer. Romance languages, which share much of the script and some vocabulary, are well separated in the last layer, but less so in the embeddings layer. Taiwanese and mainland Chinese (zh-tw and zh-cn, respectively) are well separable in the last layer, but not in the embedding layer. These findings suggest that the way mBERT encodes language identity differs across layers: while lower layers focus on lexical dimensions -- and thus cluster the two Chinese variants, and the Romance languages, together -- higher layers separate them, possibly by subtler cues, such as topical differences or syntactic alternations. This aligns with \citet{SMS19} who demonstrated that mBERT representations become more language-specific along the layers.     

To quantify the influence of the projection to the language rowspace, we calculate V-measure \citep{v-measure}, which assesses the degree of clustering according to language identity. Specifically, we perform K-means clustering with the number of languages as K, and then calculate V-measure to quantify alignment between the clusters and the language identity. On the embedding layer, this measure increases from 35.5\% in the original space, to 61.8\% on the language-identity subspace; and for the last layer, from 80.5\% in the original space, to 90.35\% in the language-identity subspace, both showing improved clustering by language identity.

When projecting the representations on the null-space we get the opposite trend: less separation by language-identity. The full results of this complementary projection can be found in Section~\ref{tsne_app} in the Appendix.

\paragraph{Inducing Language-Neutral Token Predictions}
By the disentanglement hypothesis, removing the language-encoding part of the representations should render the prediction language-agnostic. To test that, we take contextualized representations of random tokens in English sentences, and look at the original masked language model (MLM) predictions over those representations. We then compare these predictions with three variations: (a) when projecting the representations themselves on the null-space of the language-identity subspace, (b) when projecting the output embedding matrix on that null-space, (c) when projecting both the representations and the output embedding matrix on the null-space.

In order to inspect the differences in predictions we get, we train a classifier\footnote{\textsc{scikit-learn} implementation \cite{sklearn} with default parameters.} that given the embedding of a word, predicts whether it is in English or not. Then, we compute the percentage of English/non-English words in the top-k predictions for each of the variants. The results are depicted in Table~\ref{tab:lang-neut}, for $k \in {1, 5, 10, 20, 50}$. As expected, when projecting both the representations and the embeddings, we get most predictions that are not in English (the results are the average over 6000 instances).

\begin{table}
    \centering
    \scalebox{0.7}{
    \begin{tabular}{l|l|l|l|l|l}
    & top-1 & top-5 & top-10 & top-20 & top-50 \\ \hline \hline
    \textsc{None} & 0.921 & 0.807 & 0.783 & 0.773 & 0.767 \\ \hline
    \textsc{INLP \textsubscript{embed}} & 0.906 & 0.584 & 0.520 & 0.477 & 0.437 \\ \hline
    \textsc{INLP \textsubscript{repr}} & 0.908 & 0.550 & 0.484 & 0.441 & 0.403 \\ \hline
    \textsc{INLP \textsubscript{both}} & 0.868 & 0.488 & 0.414 & 0.366 & 0.325 \\ \hline
    \end{tabular}}
    \caption{Percentage of English words out of top-k predictions of the MLM, when performing INLP on the output embeddings, on the representations in context or on both.}
    \label{tab:lang-neut}
\end{table}

\begin{table}
    \centering
    \resizebox{0.5\columnwidth}{!}{
    \begin{tabular}{l|l}
    & avg-cos-sim   \\ \hline \hline
    \textsc{None} & 0.471  \\ \hline
    \textsc{INLP \textsubscript{embed}} & 0.443  \\ \hline
    \textsc{INLP \textsubscript{repr}} & 0.460  \\ \hline
    \textsc{INLP \textsubscript{both}} & 0.438  \\ \hline
    
    \end{tabular}
    }
    \caption{Average cosine-similarity between the original token and the top-10 MLM predictions for it, when performing INLP on the output embeddings, on the representations in context or on both.}
    \label{tab:avg-cos-sim}
\end{table}

The decrease in English predictions can be the result of noise that is introduced by the projection operation. To verify that the influence of the projection is focused on the langauge-identity, and not on the lexical-semantics content of the vectors, we employ a second evaluation that focuses on the semantic coherence of the predictions. We look at the top-10 predictions in each case, and compute the cosine-similarity between the original word in the sentence, and each prediction. We expect the average cosine-similarity to drop significantly if the new predictions are mostly noise. However, if the predictions are reasonably related to the original words, we expect to get a similar average. Since some of the predictions are not in English, we use MUSE cross-lingual embeddings for this evaluation \cite{CLR17}. The results are shown in Table~\ref{tab:avg-cos-sim}. As expected, the average cosine similarity is almost the same in all cases (the average is taken across the same 6000 instances). To get a sense of the resulting predictions, we show four examples (of different POS) in Table~\ref{tab:examples}. In all cases most words that were removed from the top-10 predictions are English words, while most new words are translations of the original word into other languages.

\begin{table*}
    \centering
    \resizebox{\textwidth}{!}{
    \begin{tabular}{l|l||l|l||l|l||l|l}
    \multicolumn{2}{c||}{mother} & \multicolumn{2}{c||}{sometimes} & \multicolumn{2}{c||}{visited} & \multicolumn{2}{c}{beginning}  \\ \hline \hline
    before & after & before & after & before & after & before & after \\ \hline
    mother & mother & sometimes & sometimes & visited & visited & beginning & beginning \\
    \bad{father} & moeder & \bad{soms} & manchmal & visits & visito & \bad{begin} & \selectlanguage{russian} начало \\
    madre & mothers & \selectlanguage{russian} иногда & \selectlanguage{russian} иногда & \bad{attended} & \good{besøkt} & \bad{start} & \good{{\selectlanguage{russian} початок}} \\
    mutter & \good{\selectlanguage{russian} мать} & manchmal & \good{ocasionalmente} & \bad{visit} & visits & beginn & entamu \\
    \bad{native} & \good{matki} & \bad{occasionally} & \good{ponekad} & visiting & \good{besuchte} & \bad{end} & zacatku \\
    moeder & \good{\selectlanguage{greek} μητερα} & \bad{often} & \selectlanguage{russian} често & visito & entered & \selectlanguage{russian} начало & 
    \good{\selectlanguage{arabic} \textRL{اغاز } }\\
    \bad{mary} & mutter & parfois & talvolta & entered & visiting & zacatku & \good{\selectlanguage{greek} αρχη} \\
    \bad{true} & madre & talvolta & kartais & \bad{joined} & \good{asked} & entamu & \good{pocetku} \\
    mothers & \good{\selectlanguage{arabic} \textRL{ جنس }}& \selectlanguage{russian} често & parfois & \bad{toured} & \good{vitja} & \bad{introduction} & beginn \\
    \bad{the} & \good{\selectlanguage{arabic} \textRL{ مادر }} & kartais & \good{\selectlanguage{arabic} \textRL{احيانا }} & \bad{visite} & \good{\selectlanguage{russian} посет} & \bad{comencament} & \good{zaciatku} \\
    
    \end{tabular}}
    \caption{Examples of resulting top-10 MLM predictions before and after performing INLP on both the output embeddings and representations in context. Words in red (italic) appear only in the ``before" list, while words in blue (underlined) appear only in the ``after" list.}
    \label{tab:examples}
\end{table*}

\section{Analogies-based Translation}
\label{analogies}

In the previous section we established the assumption that mBERT representations are composed of a language-neutral and a language-encoding components. In this section, we present another mechanism for word-translation with mBERT, which is based on manipulating the language-encoding component of the representation, in a similar way to how analogies in word embeddings work \cite{MYZ13}. This new method has a clear mechanism behind it, and it serves as an additional validation for our assumption about the two independent components.

The idea is simple: we create a single vector representation for each language, as explained below. Then, in order to change the embedding of a word in language SOURCE to language TARGET, we simply subtract from it the vector representation of language SOURCE and add the vector representation of language TARGET. Finally, in order to get the translation of the source word into the target language, we multiply the resulting representation by the output embedding matrix to get the closest words to it out of the full vocabulary. Below is a detailed explanation of the implementation.

\subsection{Creating language-representation vectors}

We start by extracting sentences in each language. From each sentence, we choose a random token and extract its representation from the output embedding matrix. Then, for each language we average all the obtained representations, to create a single vector representing that language. For that we use the same representations extracted for training INLP, as described in Section~\ref{inlp}. Note that no hyper-parameter tuning was done when calculating these language vectors. The assumption here is that when averaging this way, the lexical differences between the representations cancel out, while the shared language component in all of them persists.

\subsection{Performing Translation with Analogies}

We are interested in translating words from a SOURCE language to a TARGET language. For that we simply take the word embedding of the SOURCE word, subtract the representation of the SOURCE language from it, and add the representation of the TARGET language. We multiply this new representation by the output embedding matrix to get a ranking over all the vocabulary, from the closest word to it to the least close.

\subsection{Results}

In Table~\ref{tab:acc} we report the results of translation using analogies (second row). The success of this method supports the reasoning behind it -- indeed changing the language component of the representation enables us to get satisfactory results in word-level translation. While the template-based method, which is non-linear, puts a competitive lower bound on the amount of parallel information embedded in mBERT, this strictly linear method is able to recover a large portion of it.

\paragraph{Visualization of the Representation Space}

In contrast to the template-based method, t-SNE visualization of the analogies-based translation vectors reveals low clustering by language (see Figure~\ref{fig:tsne-both} in Section~\ref{vis_repr_app} in the Appendix).

\paragraph{Translation between every Language Pair}

\begin{figure}[!h]
    \centering
    \includegraphics[max width=\columnwidth]{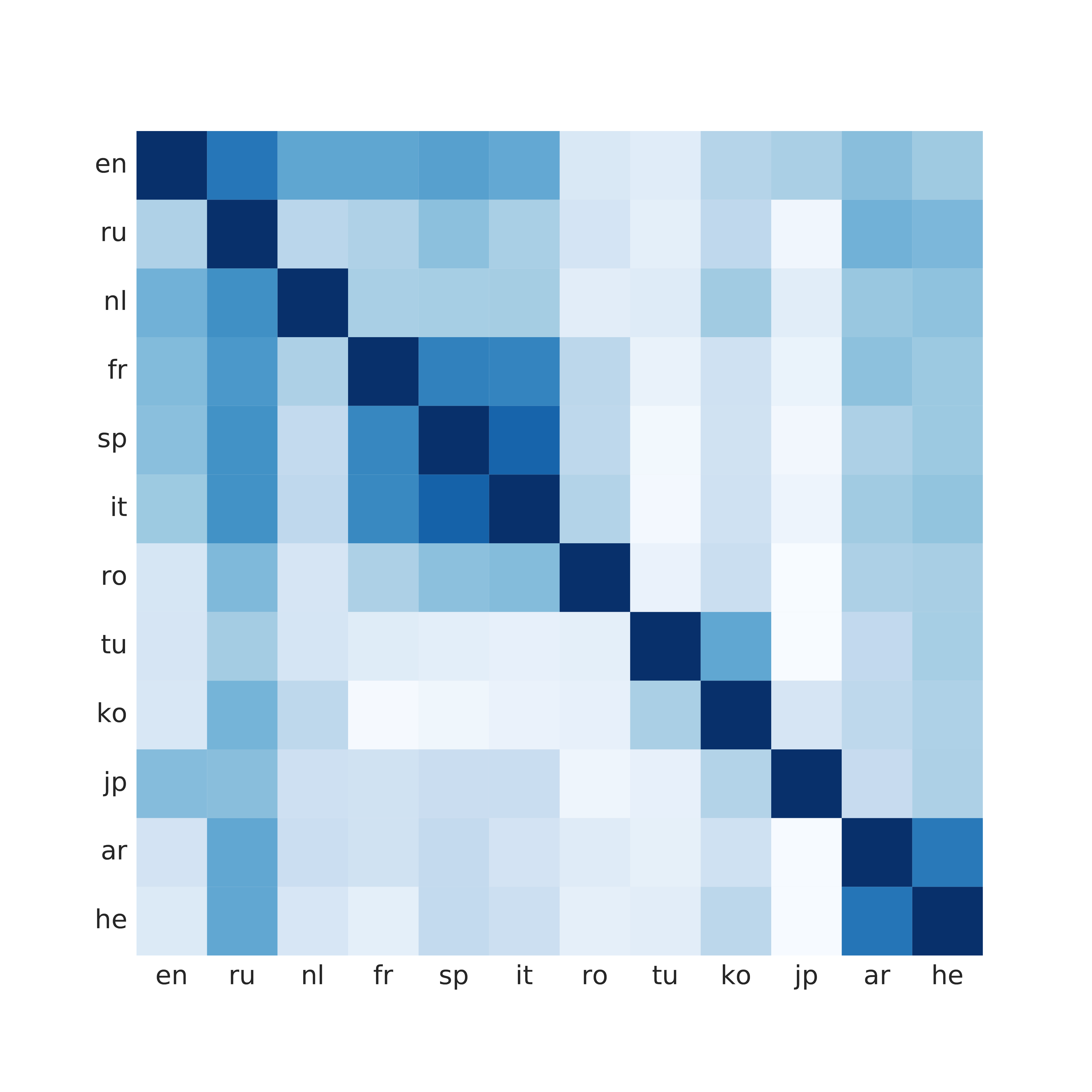}
    \caption{Acc@10 of word-level translation using analogies-based method, with source languages on the rows, and target languages on the columns.
    }
    \label{fig:heatmap_all_langs}
\end{figure}

The analogies-based translation method can be easily applied to all language pairs, by subtracting the representation vector of the source language and adding that of the target language. Figure~\ref{fig:heatmap_all_langs} presents a heatmap of the acc@10 for every language pair, with source languages on the left and target languages at the bottom. We note the high translation scores between related languages, for example, Arabic and Hebrew (both ways), and French, Spanish and Italian (all pairs).

\section{Discussion}

The template-based method we presented is non-linear and puts a high lower bound on the amount of parallel information found in mBERT, with surprisingly good results on word-level translation. The analogies-based method also gets impressive results, but to a lesser extent than the template-based one. In addition, the resulting representations in the analogies-based method are much less structured. These together suggest that most of the parallel information is not linearly decodable from mBERT.

The reasoning behind the analogies-based method is very clear: 
under linearity assumption we explicitly characterize and compute the decomposition to language-encoding and language-neutral components, and derive a word-level translation method based on this decomposition. 

The mechanism behind the template-based method and the source of its success, however, are much harder to understand and interpret. While it is possible that some parallel data, in one form or another, is present in the training corpora, this is still an implicit signal: there is no explicit supervision for the learning of translation. The fact that MLM training is sufficient -- at least to some degree -- to induce learning of the algorithmic function of translation (without further supervised finetuning) is nontrivial. We believe that the success of this method is far from being obvious. We leave further investigation of the sources of this success to future work.

\section{Conclusion}

We aim to shed light on a basic question regarding multilingual BERT: How much word-level translation information does it embed and what are the ways to extract it? answering this question can help understand the empirical findings on its impressive transfer ability across languages. 

We show that the knowledge needed for word-level translation is implicitly encoded in the model, and is easy to extract with simple methods, without fine-tuning.
This information is likely stored in a non-linear way. However, some parts of this representations can be recovered linearly: we identify an empirical language-identity subspace in mBERT, and show that under linearity assumptions, the representations in different languages are easily separable in that subspace; neutralizing the language-identity subspace encourages the model to perform word predictions which are less sensitive to language-identity, but are nonetheless semantically-meaningful. We argue that the results of those interventions support the hypothesis on the existence of identifiable language components in mBERT.

\section*{Acknowledgements}

This project has received funding from the European Research Council (ERC) under the European Union’s Horizon 2020 research and innovation programme, grant agreement No. 802774 (iEXTRACT), and from the the Israeli ministry of Science, Technology  and  Space  through  the  Israeli-French  Maimonide Cooperation programme.

\bibliography{emnlp2020}

\begin{thebibliography}{23}
\expandafter\ifx\csname natexlab\endcsname\relax\def\natexlab#1{#1}\fi

\bibitem[{Cao et~al.(2020)Cao, Kitaev, and Klein}]{CKK20}
Steven Cao, Nikita Kitaev, and Dan Klein. 2020.
\newblock Multilingual alignment of contextual word representations.
\newblock \emph{arXiv:2002.03518}.

\bibitem[{Chi et~al.(2020)Chi, Hewitt, and Manning}]{mbert-syntax}
Ethan~A. Chi, John Hewitt, and Christopher~D. Manning. 2020.
\newblock \href {http://arxiv.org/abs/2005.04511} {Finding universal
  grammatical relations in multilingual {BERT}}.
\newblock \emph{CoRR}, abs/2005.04511.

\bibitem[{Conneau et~al.(2017)Conneau, Lample, Ranzato, Denoyer, and
  J{\'e}gou}]{CLR17}
Alexis Conneau, Guillaume Lample, Marc'Aurelio Ranzato, Ludovic Denoyer, and
  Herv{\'e} J{\'e}gou. 2017.
\newblock Word translation without parallel data.
\newblock \emph{arXiv:1710.04087}.

\bibitem[{Dellert et~al.(2019)Dellert, Daneyko, M{\"u}nch, Ladygina, Buch,
  Clarius, Grigorjew, Balabel, Boga, Baysarova et~al.}]{DDM19}
Johannes Dellert, Thora Daneyko, Alla M{\"u}nch, Alina Ladygina, Armin Buch,
  Natalie Clarius, Ilja Grigorjew, Mohamed Balabel, Hizniye~Isabella Boga,
  Zalina Baysarova, et~al. 2019.
\newblock Northeuralex: a wide-coverage lexical database of northern eurasia.
\newblock \emph{Language Resources and Evaluation}, pages 1--29.

\bibitem[{Devlin et~al.(2019)Devlin, Chang, Lee, and Toutanova}]{DCL19}
Jacob Devlin, Ming-Wei Chang, Kenton Lee, and Kristina Toutanova. 2019.
\newblock \href {https://doi.org/10.18653/v1/N19-1423} {{BERT}: Pre-training of
  deep bidirectional transformers for language understanding}.
\newblock In \emph{Proceedings of the 2019 Conference of the North {A}merican
  Chapter of the Association for Computational Linguistics: Human Language
  Technologies, Volume 1 (Long and Short Papers)}, pages 4171--4186,
  Minneapolis, Minnesota. Association for Computational Linguistics.

\bibitem[{Jiang et~al.(2019)Jiang, Xu, Araki, and Neubig}]{jiang2019can}
Zhengbao Jiang, Frank~F Xu, Jun Araki, and Graham Neubig. 2019.
\newblock How can we know what language models know?
\newblock \emph{arXiv preprint arXiv:1911.12543}.

\bibitem[{Karthikeyan et~al.(2020)Karthikeyan, Wang, Mayhew, and Roth}]{KWM20}
K~Karthikeyan, Zihan Wang, Stephen Mayhew, and Dan Roth. 2020.
\newblock Cross-lingual ability of multilingual bert: An empirical study.
\newblock In \emph{International Conference on Learning Representations}.

\bibitem[{Libovick{\`y} et~al.(2019)Libovick{\`y}, Rosa, and Fraser}]{LRF19}
Jind{\v{r}}ich Libovick{\`y}, Rudolf Rosa, and Alexander Fraser. 2019.
\newblock How language-neutral is multilingual bert?
\newblock \emph{arXiv:1911.03310}.

\bibitem[{Maaten and Hinton(2008)}]{tsne}
Laurens van~der Maaten and Geoffrey Hinton. 2008.
\newblock Visualizing data using t-sne.
\newblock \emph{Journal of machine learning research}, 9(Nov):2579--2605.

\bibitem[{Mikolov et~al.(2013{\natexlab{a}})Mikolov, Chen, Corrado, and
  Dean}]{mikolov2013}
Tomas Mikolov, Kai Chen, Greg Corrado, and Jeffrey Dean. 2013{\natexlab{a}}.
\newblock Efficient estimation of word representations in vector space.
\newblock In \emph{1st International Conference on Learning Representations,
  {ICLR} 2013, Scottsdale, Arizona, USA, May 2-4, 2013, Workshop Track
  Proceedings}.

\bibitem[{Mikolov et~al.(2013{\natexlab{b}})Mikolov, Yih, and Zweig}]{MYZ13}
Tom{\'a}{\v{s}} Mikolov, Wen-tau Yih, and Geoffrey Zweig. 2013{\natexlab{b}}.
\newblock Linguistic regularities in continuous space word representations.
\newblock In \emph{Proceedings of the 2013 conference of the north american
  chapter of the association for computational linguistics: Human language
  technologies}, pages 746--751.

\bibitem[{Muller et~al.(2020)Muller, Sagot, and Seddah}]{MSS20}
Benjamin Muller, Benoˆıt Sagot, and Djame Seddah. 2020.
\newblock Can multilingual language models transfer to an unseen dialect? a
  case study on north african arabizi.
\newblock \emph{arXiv:2005.00318}.

\bibitem[{Pedregosa et~al.(2011)Pedregosa, Varoquaux, Gramfort, Michel,
  Thirion, Grisel, Blondel, Prettenhofer, Weiss, Dubourg, Vanderplas, Passos,
  Cournapeau, Brucher, Perrot, and Duchesnay}]{sklearn}
F.~Pedregosa, G.~Varoquaux, A.~Gramfort, V.~Michel, B.~Thirion, O.~Grisel,
  M.~Blondel, P.~Prettenhofer, R.~Weiss, V.~Dubourg, J.~Vanderplas, A.~Passos,
  D.~Cournapeau, M.~Brucher, M.~Perrot, and E.~Duchesnay. 2011.
\newblock Scikit-learn: Machine learning in {P}ython.
\newblock \emph{Journal of Machine Learning Research}, 12:2825--2830.

\bibitem[{Petroni et~al.(2019)Petroni, Rockt{\"a}schel, Riedel, Lewis, Bakhtin,
  Wu, and Miller}]{petroni2019language}
Fabio Petroni, Tim Rockt{\"a}schel, Sebastian Riedel, Patrick Lewis, Anton
  Bakhtin, Yuxiang Wu, and Alexander Miller. 2019.
\newblock Language models as knowledge bases?
\newblock In \emph{Proceedings of the 2019 Conference on Empirical Methods in
  Natural Language Processing and the 9th International Joint Conference on
  Natural Language Processing (EMNLP-IJCNLP)}, pages 2463--2473.

\bibitem[{Pires et~al.(2019)Pires, Schlinger, and Garrette}]{PSG19}
Telmo Pires, Eva Schlinger, and Dan Garrette. 2019.
\newblock \href {https://doi.org/10.18653/v1/P19-1493} {How multilingual is
  multilingual {BERT}?}
\newblock In \emph{Proceedings of the 57th Annual Meeting of the Association
  for Computational Linguistics}, pages 4996--5001, Florence, Italy.
  Association for Computational Linguistics.

\bibitem[{Ravfogel et~al.(2020)Ravfogel, Elazar, Gonen, Twiton, and
  Goldberg}]{inlp}
Shauli Ravfogel, Yanai Elazar, Hila Gonen, Michael Twiton, and Yoav Goldberg.
  2020.
\newblock \href {http://arxiv.org/abs/2004.07667} {Null it out: Guarding
  protected attributes by iterative nullspace projection}.
\newblock \emph{CoRR}, abs/2004.07667.

\bibitem[{Rosenberg and Hirschberg(2007)}]{v-measure}
Andrew Rosenberg and Julia Hirschberg. 2007.
\newblock \href {https://www.aclweb.org/anthology/D07-1043/} {V-measure: {A}
  conditional entropy-based external cluster evaluation measure}.
\newblock In \emph{EMNLP-CoNLL 2007, Proceedings of the 2007 Joint Conference
  on Empirical Methods in Natural Language Processing and Computational Natural
  Language Learning, June 28-30, 2007, Prague, Czech Republic}, pages 410--420.
  {ACL}.

\bibitem[{Singh et~al.(2019)Singh, McCann, Socher, and Xiong}]{SMS19}
Jasdeep Singh, Bryan McCann, Richard Socher, and Caiming Xiong. 2019.
\newblock \href {https://doi.org/10.18653/v1/D19-6106} {{BERT} is not an
  interlingua and the bias of tokenization}.
\newblock In \emph{Proceedings of the 2nd Workshop on Deep Learning Approaches
  for Low-Resource NLP (DeepLo 2019)}, pages 47--55, Hong Kong, China.
  Association for Computational Linguistics.

\bibitem[{Talmor et~al.(2019)Talmor, Elazar, Goldberg, and
  Berant}]{talmor2019olmpics}
Alon Talmor, Yanai Elazar, Yoav Goldberg, and Jonathan Berant. 2019.
\newblock \href {http://arxiv.org/abs/1912.13283} {olmpics -- on what language
  model pre-training captures}.

\bibitem[{Wang et~al.(2019)Wang, Che, Guo, Liu, and Liu}]{WCG19}
Yuxuan Wang, Wanxiang Che, Jiang Guo, Yijia Liu, and Ting Liu. 2019.
\newblock \href {https://doi.org/10.18653/v1/D19-1575} {Cross-lingual {BERT}
  transformation for zero-shot dependency parsing}.
\newblock In \emph{Proceedings of the 2019 Conference on Empirical Methods in
  Natural Language Processing and the 9th International Joint Conference on
  Natural Language Processing (EMNLP-IJCNLP)}, pages 5721--5727, Hong Kong,
  China. Association for Computational Linguistics.

\bibitem[{Wolf et~al.(2019)Wolf, Debut, Sanh, Chaumond, Delangue, Moi, Cistac,
  Rault, Louf, Funtowicz, Davison, Shleifer, von Platen, Ma, Jernite, Plu, Xu,
  Scao, Gugger, Drame, Lhoest, and Rush}]{trans_hf}
Thomas Wolf, Lysandre Debut, Victor Sanh, Julien Chaumond, Clement Delangue,
  Anthony Moi, Pierric Cistac, Tim Rault, Rémi Louf, Morgan Funtowicz, Joe
  Davison, Sam Shleifer, Patrick von Platen, Clara Ma, Yacine Jernite, Julien
  Plu, Canwen Xu, Teven~Le Scao, Sylvain Gugger, Mariama Drame, Quentin Lhoest,
  and Alexander~M. Rush. 2019.
\newblock Huggingface's transformers: State-of-the-art natural language
  processing.
\newblock \emph{ArXiv}, abs/1910.03771.

\bibitem[{Wu and Dredze(2019)}]{WD19}
Shijie Wu and Mark Dredze. 2019.
\newblock \href {https://doi.org/10.18653/v1/D19-1077} {Beto, bentz, becas: The
  surprising cross-lingual effectiveness of {BERT}}.
\newblock In \emph{Proceedings of the 2019 Conference on Empirical Methods in
  Natural Language Processing and the 9th International Joint Conference on
  Natural Language Processing (EMNLP-IJCNLP)}, pages 833--844, Hong Kong,
  China. Association for Computational Linguistics.

\bibitem[{Ye et~al.(2018)Ye, Devendra, Matthieu, Sarguna, and Graham}]{ted}
Qi~Ye, Sachan Devendra, Felix Matthieu, Padmanabhan Sarguna, and Neubig Graham.
  2018.
\newblock When and why are pre-trained word embeddings useful for neural
  machine translation.
\newblock In \emph{HLT-NAACL}.

\end{thebibliography}
\bibliographystyle{acl_natbib}

\clearpage

\appendix

\section{Templates}
\label{templates_app}

The different templates are listed in Table~\ref{tab:templates}, from the best performing to least performing. We report the results throughout the paper using the best template (first). Templates 5--7 fail completely, while templates 2--4 result in reasonable accuracy.

\begin{table}[h!]
    \centering
    \scalebox{0.58}{
    \begin{tabular}{l|l}
    1 & ``The word `\textsc{source}' in \textsc{language} is: [MASK]." \\ \hline
    2 & ```\textsc{source}' in \textsc{language} is: [MASK]." \\ \hline
    3 & ``Translate the word `\textsc{source}' into \textsc{language}: [MASK]." \\ \hline
    4 & ``What is the meaning of the \textsc{language} word [MASK]? `\textsc{source}'." \\ \hline
    5 & ``What is the translation of the word `\textsc{source}' into \textsc{language}? [MASK]." \\ \hline
    6 & ``The translation of the word `\textsc{source}' into \textsc{language} is [MASK]." \\ \hline
    7 & ``How do you say `\textsc{source}' in \textsc{language}? [MASK]." \\ \hline
    \end{tabular}}
    \caption{The different templates we experimented with.}
    \label{tab:templates}
\end{table}

\section{Predicting the Language}

Table~\ref{tab:lang_guess_app} depicts the results of language prediction from the template. We report acc@1,5,10 for all languages.

\begin{table}
    \centering
    \begin{tabular}{lrrr}
    \toprule
    languages &  acc@1 &  acc@5 &  acc@10 \\
    \midrule
    greek &  0.986 &  0.998 &  1.000 \\
    russian &  0.943 &  0.994 &  0.998 \\
    arabic &  0.794 &  0.963 &  0.984 \\
    hebrew &  0.761 &  0.968 &  0.990 \\
    german &  0.758 &  0.951 &  0.991 \\
    japanese &  0.716 &  0.939 &  0.966 \\
    korean &  0.664 &  0.905 &  0.949 \\
    french &  0.637 &  0.976 &  0.992 \\
    latin &  0.626 &  0.900 &  0.959 \\
    polish &  0.576 &  0.728 &  0.803 \\
    italian &  0.572 &  0.873 &  0.947 \\
    spanish &  0.503 &  0.757 &  0.878 \\
    finnish &  0.404 &  0.622 &  0.748 \\
    turkish &  0.399 &  0.589 &  0.709 \\
    dutch &  0.315 &  0.846 &  0.965 \\
    welsh &  0.262 &  0.548 &  0.680 \\
    swedish &  0.262 &  0.492 &  0.692 \\
    hungarian &  0.254 &  0.395 &  0.493 \\
    portuguese &  0.236 &  0.587 &  0.808 \\
    danish &  0.231 &  0.388 &  0.567 \\
    malayalam &  0.216 &  0.587 &  0.712 \\
    irish &  0.182 &  0.598 &  0.762 \\
    kannada &  0.180 &  0.714 &  0.850 \\
    lithuanian &  0.170 &  0.357 &  0.507 \\
    bengali &  0.156 &  0.609 &  0.746 \\
    ukrainian &  0.147 &  0.831 &  0.894 \\
    telugu &  0.132 &  0.636 &  0.819 \\
    basque &  0.126 &  0.249 &  0.359 \\
    estonian &  0.126 &  0.374 &  0.510 \\
    albanian &  0.117 &  0.225 &  0.323 \\
    croatian &  0.106 &  0.279 &  0.353 \\
    norwegian &  0.105 &  0.280 &  0.395 \\
    catalan &  0.104 &  0.317 &  0.512 \\
    czech &  0.080 &  0.251 &  0.361 \\
    romanian &  0.069 &  0.110 &  0.164 \\
    armenian &  0.060 &  0.368 &  0.588 \\
    tamil &  0.057 &  0.467 &  0.608 \\
    bulgarian &  0.040 &  0.582 &  0.815 \\
    hindi &  0.037 &  0.467 &  0.775 \\
    latvian &  0.036 &  0.118 &  0.202 \\
    breton &  0.017 &  0.127 &  0.209 \\
    slovak &  0.003 &  0.076 &  0.196 \\
    farsi &  0.000 &  0.001 &  0.013 \\
    tatar &  0.000 &  0.008 &  0.092 \\
    ket &  0.000 &  0.000 &  0.000 \\
    georgian &  0.000 &  0.073 &  0.183 \\
    lak &  0.000 &  0.000 &  0.000 \\
    \bottomrule
    \end{tabular}
    \caption{Prediction accuracy of the language, when the language is masked in the template.}
    \label{tab:lang_guess_app}
\end{table}

\section{t-SNE for Nullsapce-Projected Vectors}
\label{tsne_app}

\begin{figure}[h!]
\subfigure{
\includegraphics[width=\columnwidth]{contextualized.original.pdf}
}
\subfigure{
\includegraphics[width=\columnwidth]{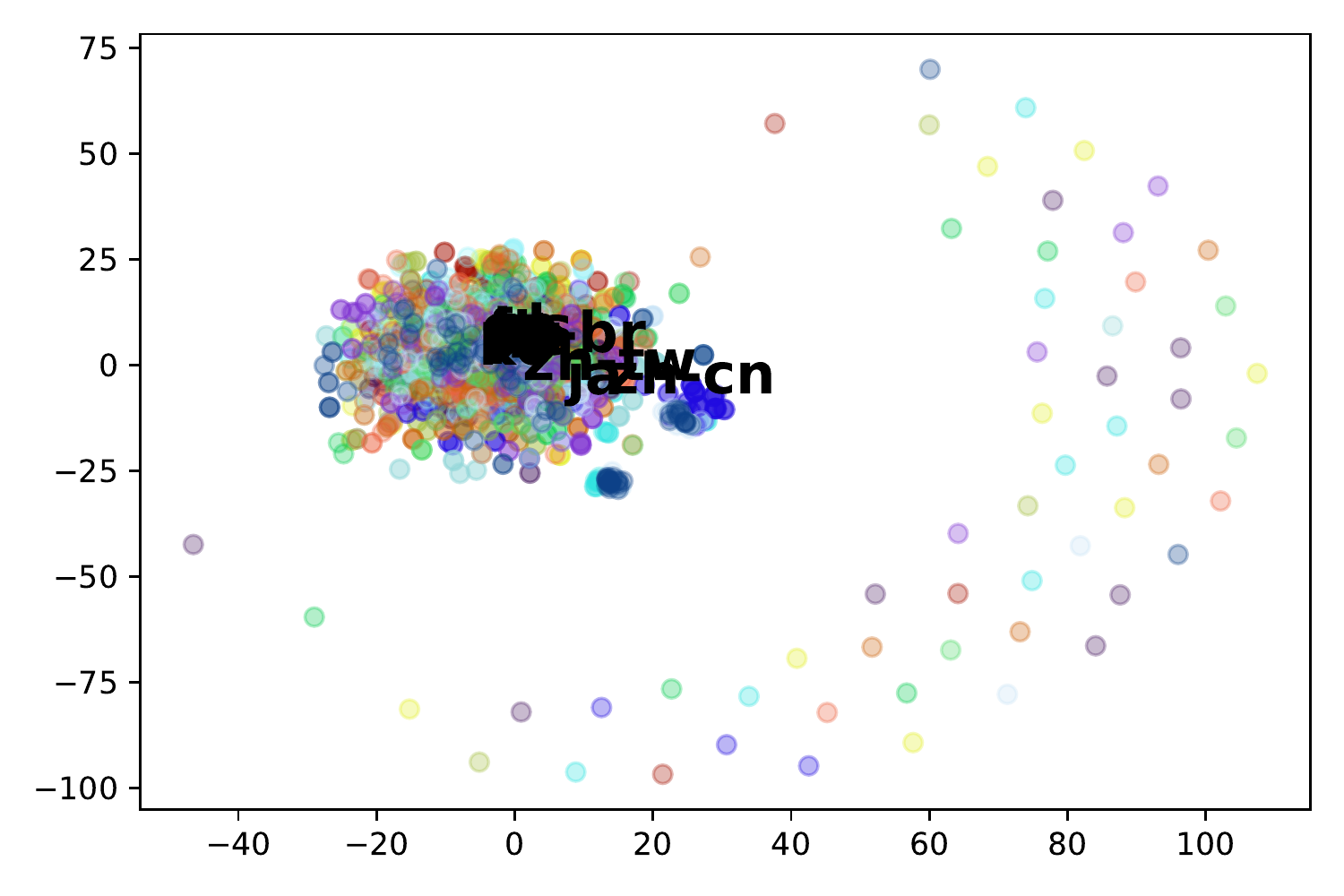}
}
\caption{t-SNE projection of the last-hidden-layer representations of random words from different language, originally (top) and after (bottom) nullspace projection.}
\label{fig:tsne-original-vs-nullspace:layer12}
\end{figure}

\begin{figure}[h!]
\subfigure{
\includegraphics[width=\columnwidth]{embeddings.original.pdf}
}
\subfigure{
\includegraphics[width=\columnwidth]{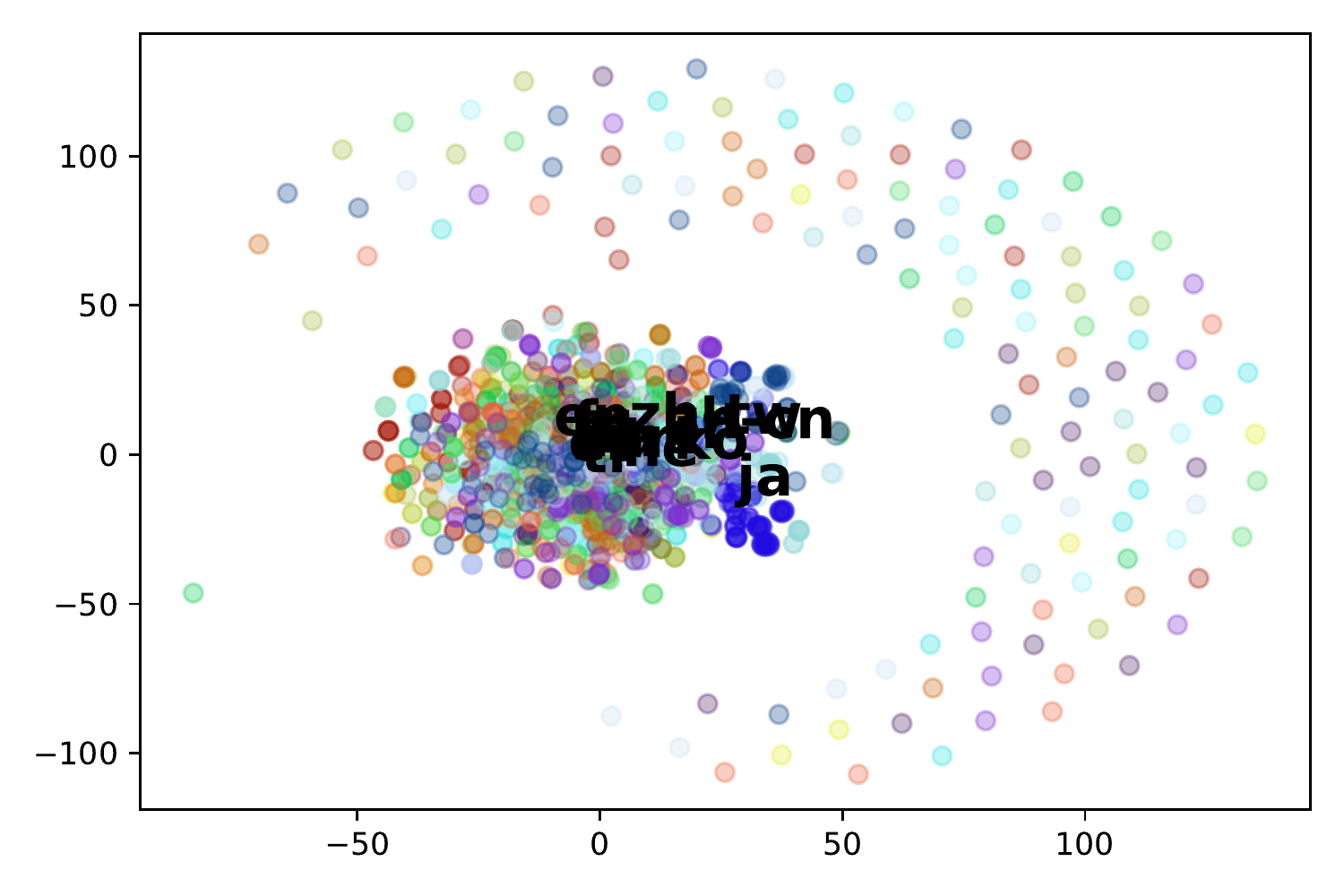}
}
\caption{t-SNE projection of the output embeddings of random words from different language, originally (top) and after (bottom) nullspace projection.}
\label{fig:tsne-original-vs-nullspace:layer0}
\end{figure}

In figures \ref{fig:tsne-original-vs-nullspace:layer12} and \ref{fig:tsne-original-vs-nullspace:layer0} we present a t-SNE projection of the representations in the embeddings layer and the last layer, projected onto INLP nullsapce -- a subspace which discards information relevant for language-identity prediction. As expected, the nullspace does not encode language identity: V-measure drops to 11.5\% and 11.4\% in the embeddings layer and in the last layer, respectively.

\section{Visualization of the Representation Space}
\label{vis_repr_app}
\begin{figure}[h!]
	\centering
	\scalebox{0.3}{
	\includegraphics{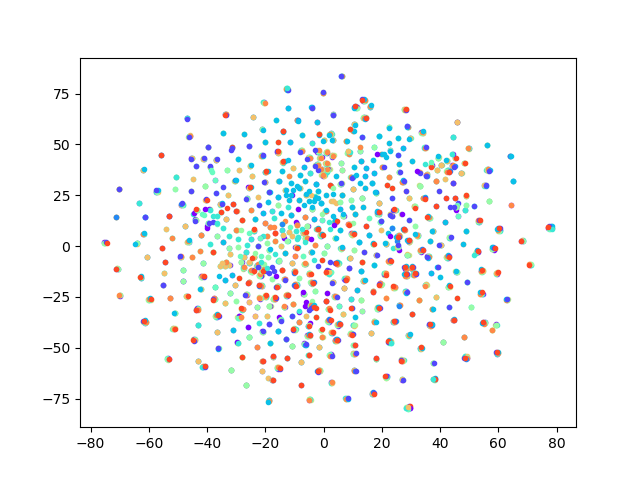}}
	\caption{t-SNE projections of the representations of the analogies-based method.}
	\label{fig:tsne-both}
\end{figure}

We plot the t-SNE projection of the representations of the analogies-based method (after subtraction and addition of the language vectors), colored by target language. While the representations of the template-based method are clearly clustered according to the target language, the representations in this method are completely mixed, see Figure~\ref{fig:tsne-both}. 

\end{document}